\documentclass{article}
\usepackage{spconf,amsmath,graphicx}

\usepackage{multirow}
\usepackage{amsmath}
\usepackage{booktabs}
\usepackage[export]{adjustbox}
\usepackage{tabularx}
\usepackage{hyperref}
\usepackage{xcolor}

\title{ST360IQ: No-Reference Omnidirectional Image Quality Assessment with Spherical Vision Transformers}

\name{\begin{tabular}{c}Nafiseh Jabbari Tofighi$^\star$, Mohamed Hedi Elfkir$^\dagger$, Nevrez Imamoglu$^\ast$, Cagri Ozcinar$^\ddagger$\\
\textit{Erkut Erdem}$^\dagger$, \textit{Aykut Erdem}$^\star$
\end{tabular}}
\address{$^\star$ Ko\c{c} University, KUIS AI Center$\quad^\dagger$ Hacettepe University$\quad^\ast$ AIST Japan$\quad^\ddagger$ MSK.AI}

\begin{document}

\maketitle

%

\begin{abstract}

Omnidirectional images, aka 360$^\circ$ images, can deliver immersive and interactive visual experiences. As their popularity has increased dramatically in recent years, evaluating the quality of 360$^\circ$ images has become a problem of interest since it provides insights for capturing, transmitting, and consuming this new media. However, directly adapting quality assessment methods proposed for standard natural images for omnidirectional data poses certain challenges. These models need to deal with very high-resolution data and implicit distortions due to the spherical form of the images. In this study, we present a  method for no-reference 360$^\circ$ image quality assessment. Our proposed ST360IQ model extracts tangent viewports from the salient parts of the input omnidirectional image and employs a vision-transformers based module processing saliency selective patches/tokens that estimates a quality score from each viewport. Then, it aggregates these scores to give a final quality score. Our experiments on two benchmark datasets, namely OIQA and CVIQ datasets, demonstrate that as compared to the state-of-the-art, our approach predicts the quality of an omnidirectional image correlated with the human-perceived image quality. The code has been available on
\color{magenta}\small{\url{https://github.com/Nafiseh-Tofighi/ST360IQ}}
\end{abstract}
\begin{keywords}
Vision Transformers, 360° image quality assessment
\end{keywords}
\begin{figure*}[htb]
  \centering
  \centerline{\includegraphics[width=0.9\linewidth]{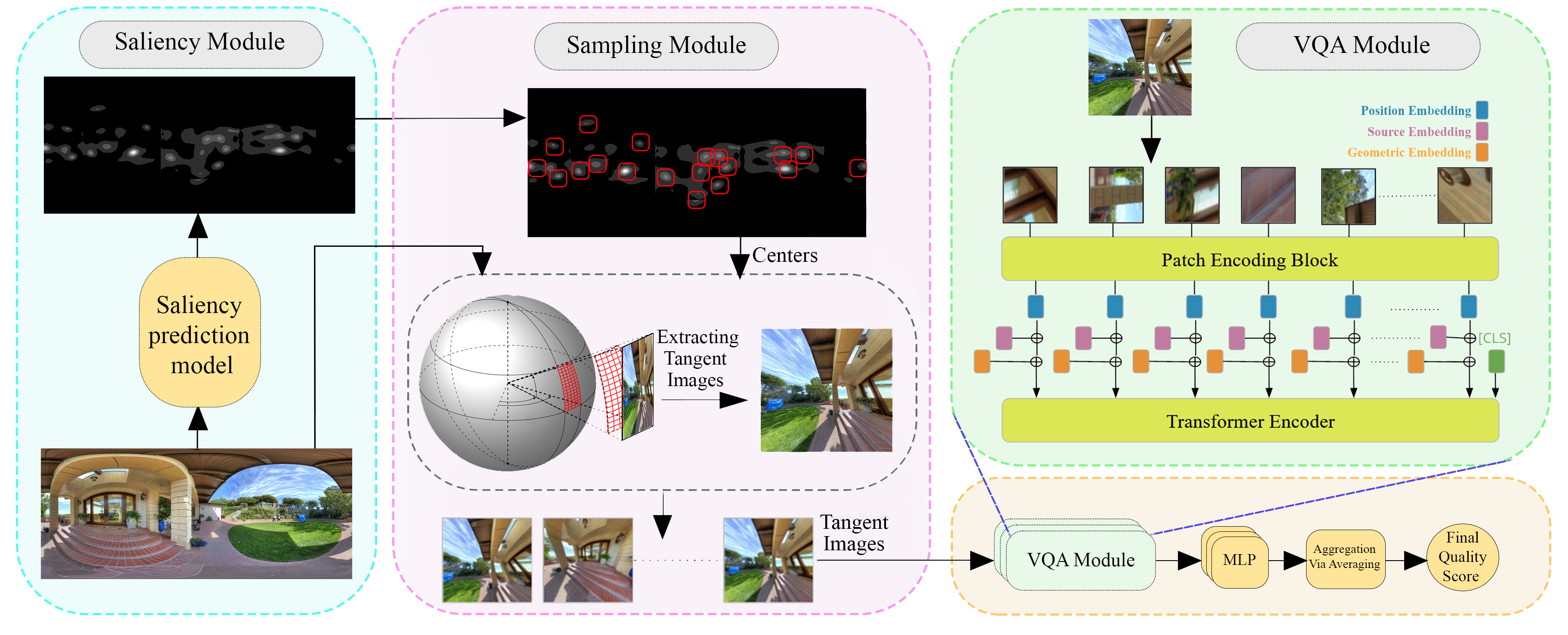}}
\caption{Overview of the our ST360IQ model for 360$^\circ$ image quality assessment, which employs a novel saliency-driven sampling. Our proposed ViT-based model processes each one of these extracted viewports by converting them into a series of tokens and then adding positional, geometric and source embeddings to better capture omnidirectional information. A MLP layer transforms the encoder’s output to a scalar value. Finally, we aggregate these scores to get the final quality score.}
\label{fig:res}
\end{figure*}
\section{Introduction}
\label{sec:intro}

Virtual reality (VR) data has recently occupied an increasing share of multimedia data as the corresponding hardware, including Head Mounted Displays (HMD) and capture devices, has become more widely accessible to the end-users.
Therefore, developing methods for evaluating the quality of omnidirectional images has also gained much attention from the researchers working in this area.

Image quality assessment (IQA) has been thoroughly studied in the past twenty years~\cite{DBLP:journals/corr/abs-2004-07728}. Generally speaking, IQA algorithms can be classified into full-reference IQA (FR IQA), reduced-reference IQA (RR IQA), and no-reference IQA (NR IQA)~\cite{Akhtar2017}. FR IQA and RR IQA models need full and part reference image information, respectively,  while NR IQA takes only the distorted image as input. In that sense, NR IQA measures subjective, perceptual quality of images.

With the recent advances in deep learning, deep methods started to dominate the IQA literature. While the early examples employ convolutionals neural networks (CNNs) for predicting the image quality~\cite{DBLP:journals/corr/abs-2004-07728,DBLP:journals/corr/abs-1801-03924}, the more recent work considers self-attention and vision-transformers (ViT)~\cite{dosovitskiy2020image} based architectures~\cite{you2021transformer,ke2021musiq}. In general, the quality assessment literature for 360$^\circ$ images follows the IQA literature as most of the existing work either suggest to directly use the same IQA methods proposed for natural images, or make small modification to those methods to deal with the spherical form~\cite{xu2020state}. Only very recently, CNN-based models specific to omnidirection images haves been proposed~\cite{8702664,xu2020blind,9506044,9766332,9859468}.

In our work, we propose a ViT-based framework for omnidirectional image quality assessment called \emph{ST360IQ}. Up to our knowledge, there are no prior work that explores ViTs for evaluating the perceptual quality of 360$^\circ$ images. Our framework estimates the overall image quality through a number of tangent viewports focusing on different parts of the spherical image. In particular, motivated by the relation between visual salience and image quality~\cite{zhang2014vsi}, we sample these viewports from the salient parts of the omnidirectional data. We process each viewport independently through a spherical ViT-based module, and extract individual quality scores, which are then aggregated to obtain the final quality score. Using tangent viewports instead of the commonly used projections such as equirectangular projection (ERP) or cube-mapping allows us to deal with  distorted visual data due to the spherical form more effectively. Moreover, our ViT-based formulation specifically designed for the omnidirectional data domain results in more accurate predictions than the state-of-the-art. 


\section{Proposed Method}
\label{sec:format}
To address the challenge of learning to assess the quality of omnidirectional images, we propose a ViT-based method that utilizes a saliency-driven sampling technique to better capture important parts of panoramic content. Figure 1 shows a general overview of our model. In the saliency module, we first predict visually attractive parts of the panoramic image via an off-the-shelf visual saliency prediction model. Our sampling module then extracts tangent viewports from salient regions of the distorted image in ERP format. The main VQA module resembles the one used in \cite{ke2021musiq}, but the patch encoding block is re-modeled from the ground up to deal with the special structure of 360$^\circ$ images. 
 To effectively encode input viewports into a sequence of tokens, we incorporate positional, geometric and source embeddings into the extracted sequence of tokens, and add a learnable classification token (CLS) to capture the global representation for the image. The corresponding 360° image quality score is predicted by the output of a fully connected layer on top of the final CLS token representation at the output of the Transformer encoder.

\noindent\textbf{Sampling Module.} Most 360$^\circ$ IQA datasets store images in the ERP format, which is the most popular spherical image representation, but is known to have significant distortions. To improve the model performance in 360$^\circ$ IQA task, we utilize tangent image representation \cite{eder2019convolutions} to reduce the distortion of the viewports extracted from the ERP image. Moreover, to distinguish visually important parts of a panoramic image, we designed a sampling strategy motivated by the human visual attention mechanism. In particular, we employ the 360$^\circ$ image saliency prediction model named ATSal \cite{dahou2021atsal} to predict salient regions of the panorama. In this way, one can assign a saliency score for each patch extracted from the omnidirectional image, which we utilize in our sampling module.

Our main motivation behind our saliency-guided sampling scheme is to combine neural attention (self-attention) mechanism with human visual attention in a simple and intuitive manner. At the first step of the sampling module, an input image in ERP format is fed to the ATSal saliency prediction model to predict a saliency map showing which regions are most likely to attract attention. The mean shift algorithm is then applied to the extracted saliency map to better highlight the salient regions. Finally, we randomly sample~10\% of regions wrt the mean saliency scores computed within each overlapping region extracted with a stride value of~S.  

In this work, we utilize tangent image representation for 360$^\circ$ images, which divide the sphere into several nearly Euclidean zones. Using the center points of each sampled salient region ($\Phi_n$, $\theta_n$), we construct a tangent image, which is free from the distortions observed in ERP images. In addition, this enables the model to gain more effective information from the same viewport size thanks to its wider field of view. As our model processes each tangent image separately during training, the quality scores of these tangent images are set to the ground truth quality score assigned to the whole image.

\begin{table*}[!t]
\caption{Quantitative comparison of ST360IQ against the state-of-the-art on CVIQ. Bold scores indicate the best performances.}
\centering
\resizebox{0.9\linewidth}{!}{%
\begin{tabular}{clc@{$\;\,$}c@{$\;\,$}c@{$\quad$}c@{$\;\,$}c@{$\;\,$}c@{$\quad$}c@{$\;\,$}c@{$\;\,$}c@{$\quad$}c@{$\;\,$}c@{$\;\,$}c@{$\quad$}}
\toprule
& & \multicolumn{3}{c}{JPEG} & \multicolumn{3}{c}{H.264/AVC} & \multicolumn{3}{c}{H.265/HEVC} & \multicolumn{3}{c}{Overall} \\ 
\midrule
        & Method & PLCC$\uparrow$ & SRCC$\uparrow$ & RMSE$\downarrow$ & PLCC$\uparrow$ & SRCC$\uparrow$ & RMSE$\downarrow$ & PLCC$\uparrow$ & SRCC$\uparrow$ & RMSE$\downarrow$ & PLCC$\uparrow$ & SRCC$\uparrow$ & RMSE$\downarrow$\\ 
\midrule
\parbox[t]{\tabcolsep}{\multirow{12}{*}{\rotatebox[origin=c]{90}{Full Reference}}} & PSNR & 0.75 & 0.76 & 10.66 & 0.66 & 0.66 & 10.06 & 0.60 & 0.57 & 9.47 & 0.65 & 0.68 & 10.65 \\ 
   & SSIM~\cite{Wang:2004} & 0.98 & 0.95 & 3.41 & 0.88 & 0.86 & 6.39 & 0.85 & 0.82 & 6.28 & 0.90 & 0.87 & 5.95 \\ 
   & FSIM~\cite{5705575} & 0.98 & 0.96 & 3.37 & 0.95 & 0.94 & 4.34 & 0.95 & 0.95 & 3.67 & 0.95 & 0.94 & 4.50 \\
   & MS\_SSIM~\cite{1292216} & 0.95 & 0.89 & 5.06 & 0.75 & 0.73 & 8.78 & 0.73 & 0.72 & 8.02 & 0.83 & 0.78 & 7.88 \\
   & IW\_SSIM~\cite{wang2010information} & 0.98 & 0.96 & 3.03 & 0.94 & 0.94 & 4.37 & 0.95 & 0.95 & 3.63 & 0.91 & 0.90 & 5.71 \\ 
   & SR-SIM~\cite{6467149} & 0.97 & 0.94 & 3.92 & 0.89 & 0.86 & 6.19 & 0.91 & 0.89 & 4.99 & 0.88 & 0.86 & 6.52 \\ 
   & GMSD~\cite{DBLP:journals/corr/XueZMB13} & 0.96 & 0.91 & 4.28 & 0.73 & 0.72 & 9.07 & 0.81 & 0.81 & 6.96 & 0.82 & 0.79 & 8.03 \\ 
   & VSI~\cite{zhang2014vsi} & 0.96 & 0.91 & 4.59 & 0.87 & 0.85 & 6.67 & 0.86 & 0.84 & 5.97 & 0.89 & 0.85 & 6.41 \\ 
   & HaarPSI~\cite{DBLP:journals/corr/ReisenhoferBKW16} & 0.97 & 0.95 & 3.63 & 0.87 & 0.85 & 6.55 & 0.89 & 0.88 & 5.27 & 0.90 & 0.87 & 5.98 \\ 
   & LPIPS~\cite{DBLP:journals/corr/abs-1801-03924} & 0.93 & 0.85 & 6.07 & 0.96 & 0.96 & 3.77 & 0.95 & 0.95 & 3.85 & 0.92 & 0.91 & 5.53 \\ 
   & DISTS~\cite{DBLP:journals/corr/abs-2004-07728} & 0.96 & 0.91 & 4.74 & 0.97 & 0.97 & 3.34 & 0.96 & 0.96 & 3.34 & 0.94 & 0.93 & 4.90 \\ 
   & MDSI~\cite{DBLP:journals/corr/NafchiSHC16} & 0.98 & 0.95 & 3.41 & 0.91 & 0.89 & 5.44 & 0.93 & 0.92 & 4.30 & 0.92 & 0.90 & 5.46 \\ 
   \midrule
   \parbox[t]{\tabcolsep}{\multirow{4}{*}{\rotatebox[origin=c]{90}{No Ref.}}} & BRISQUE~\cite{6272356} & 0.86 & 0.83 & 8.31 & 0.81 & 0.90 & 13.37 & 0.62 & 0.79 & 9.27 & 0.75 & 0.78 & 9.21 \\ 
   & MUSIQ~\cite{ke2021musiq} & 0.94 & 0.84 & 5.55 & 0.90 & 0.84 & 5.85 & 0.85 & 0.81 & 6.24 & 0.89 & 0.81 & 6.43 \\
   & MC360IQA~\cite{8702664} & 0.96 & 0.96 & 4.30 & 0.96 & 0.96 & 3.65 & 0.90 & 0.91 & 5.00 & 0.95 & 0.95 & 4.65 \\ 
   & VGCN~\cite{xu2020blind} & \textbf{0.99} & \textbf{0.98} & \textbf{2.50} & 0.97 & 0.97 & 3.15 & 0.94 & 0.95 & 3.99 & 0.96 & 0.96 & 3.67 \\ 
   & ST360IQ (Ours)& \textbf{0.99} & 0.97 & 2.67 & \textbf{0.99} & \textbf{0.98} & \textbf{2.06} & \textbf{0.96} & \textbf{0.96} & \textbf{3.25} & \textbf{0.98} & \textbf{0.98} & \textbf{2.98} \\ 
\bottomrule
\end{tabular}
}
\label{table:cviq}
\end{table*}

\noindent\textbf{Patch Encoder and Model Embeddings.} After the sampling module, it is necessary to generate transformer input patches from selected viewports. 
Instead of the linear projection as used in the original ViT model~\cite{dosovitskiy2020image}, we utilize ResNet-50 convolutional layers, \cite{he2016deep} is used for encoding the patches. 
The embedding information is added at the top of the output of the patch encoder part. 
Position embeddings are added to the patches to retain positional information, such as standard vision transformer architecture, through each image patch. 
The model keeps the information about each viewport extracted from which region of the original input image with ERP as a consequence of Geometric embedding.
This enables the model to learn better with the help of neighbor viewports correlation information. 
To this aim, the central data of each tangent viewport ($\Phi_n$, $\theta_n$) after normalizing are added to the patches as geometric embedding.
Last but not least, source embedding is employed since each viewport is fed into the transformer encoder independently, and with this feature, the network has the opportunity of predicting even with a single or limited number of viewports. To achieve this, all tangent viewports extracted from a single ERP input image gain the same source embedding corresponding to their ERP index. This helps the model mark each tangent image comes from which input source. As a result of this architecture, the average score of all tangent images extracted from a single ERP input is reported as the final prediction score for each image.

\textbf{Model Training and Implementation Details.} Input tokens of the vanilla Transformer used in this work have dimensions of $D = 384$, and we employ a patch size of $P = 32$. To make the model size equivalent to ResNet-50, we utilize a classic Transformer with six heads, 14 layers, 384 hidden sizes, and 1152 MLP size. For sampling, each input image is divided into regions by stride 16, and then 10\% of those salient regions are selected, which gives empirically the best results. The training pipeline uses the mean absolute error (MAE) as the loss function.

\section{Experimental Results}
\noindent\textbf{Datasets.} We evaluate the performance of our method using  the commonly used CVIQ~\cite{sun2018large} and OIQA~\cite{8351786} datasets. CVIQ dataset contains 528 compressed images that are obtained by applying three widely used coding standards (JPEG, H.264/AVC, and H.265/HEVC) on 16 lossless images. OIQA dataset respectively consists of 16 raw, and 320 distorted 360$^\circ$ images that are obtained from the raw images by applying four common distortion types (Gaussian blur, Gaussian noise, JPEG compression, and JPEG2000 compression) from five different levels. Following~\cite{li2019viewport,xu2020blind}, for our analysis, we randomly decompose the aforementioned datasets into train ($\sim$80\%) and ($\sim$20\%) test splits along the lossless/distortion-free image dimension. This results in 13 training and 3 test images for both datasets. By this way, we guarantee that the test images, apart from whether they are distorted or distortion-free, have not been seen during training.

\begin{table*}[!t]
\caption{Quantitative comparison of ST360IQ against the state-of-the-art on OIQA. Bold scores indicate the best performances.}
\centering
\resizebox{\linewidth}{!}{%
\begin{tabular}{clc@{$\;\,$}c@{$\;\,$}c@{$\quad$}c@{$\;\,$}c@{$\;\,$}c@{$\quad$}c@{$\;\,$}c@{$\;\,$}c@{$\quad$}c@{$\;\,$}c@{$\;\,$}c@{$\quad$}c@{$\;\,$}c@{$\;\,$}c}
\toprule
& & \multicolumn{3}{c}{JPEG} & \multicolumn{3}{c}{JPEG2000} & \multicolumn{3}{c}{Gaussian Blur } & \multicolumn{3}{c}{Gaussian Noise} & \multicolumn{3}{c}{Overall} \\ 
\midrule
        & Method & PLCC$\uparrow$ & SRCC$\uparrow$ & RMSE$\downarrow$ & PLCC$\uparrow$ & SRCC$\uparrow$ & RMSE$\downarrow$ & PLCC$\uparrow$ & SRCC$\uparrow$ & RMSE$\downarrow$ & PLCC$\uparrow$ & SRCC$\uparrow$ & RMSE$\downarrow$ & PLCC$\uparrow$ & SRCC$\uparrow$ & RMSE$\downarrow$ \\ 
\midrule
\parbox[t]{\tabcolsep}{\multirow{12}{*}{\rotatebox[origin=c]{90}{Full Reference}}} & PSNR & 0.75 & 0.72 & 1.43 & 0.88 & 0.89 & 1.01 & 0.97 & 0.96 & 1.73 & 0.77 & 0.82 & 1.17 & 0.64 & 0.60 & 1.54 \\ 
   & SSIM~\cite{Wang:2004} & 0.94 & 0.96 & 0.73 & 0.96 & 0.97 & 0.54 & 0.97 & 0.95 & 0.41 & 0.96 & 0.96 & 0.51 & 0.92 & 0.92 & 0.77 \\ 
   & FSIM~\cite{5705575} & 0.95 & 0.96 & 0.65 & 0.95 & 0.95 & 0.63 & 0.97 & 0.96 & 0.40 & 0.96 & 0.96 & 0.48 & 0.93 & 0.93 & 0.72 \\ 
   & MS\_SSIM~\cite{1292216} & 0.97 & 0.94 & 0.49 & 0.94 & 0.92 & 0.74 & 0.88 & 0.87 & 0.79 & 0.82 & 0.84 & 1.04 & 0.70 & 0.68 & 1.42 \\
   & IW\_SSIM~\cite{wang2010information} & 0.95 & 0.96 & 0.67 & 0.97 & 0.97 & 0.44 & 0.87 & 0.84 & 0.84 & 0.91 & 0.92 & 0.73 & 0.76 & 0.75 & 1.31 \\ 
   & SR-SIM~\cite{6467149} & 0.94 & 0.96 & 0.73 & 0.95 & 0.96 & 0.62 & 0.96 & 0.95 & 0.42 & 0.96 & 0.96 & 0.51 & 0.92 & 0.93 & 0.75 \\ 
   & GMSD~\cite{DBLP:journals/corr/XueZMB13} & 0.95 & 0.94 & 0.93 & 0.95 & 0.93 & 0.65 & 0.90 & 0.85 & 0.74 & 0.85 & 0.88 & 0.96 & 0.77 & 0.76 & 1.27 \\ 
   & VSI~\cite{zhang2014vsi} & 0.95 & 0.97 & 0.65 & 0.96 & 0.96 & 0.56 & 0.97 & 0.96 & 0.38 & 0.96 & 0.96 & 0.94 & 0.93 & 0.93 & 0.72 \\ 
   & HaarPSI~\cite{DBLP:journals/corr/ReisenhoferBKW16} & 0.95 & 0.96 & 0.65 & \textbf{0.98} & \textbf{0.97} & \textbf{0.36} & 0.91 & 0.91 & 0.70 & 0.92 & 0.91 & 0.70 & 0.84 & 0.83 & 1.077 \\ 
   & LPIPS~\cite{DBLP:journals/corr/abs-1801-03924} & 0.98 & 0.97 & 0.41 & 0.91 & 0.91 & 0.86 & 0.96 & 0.95 & 0.48 & 0.94 & 0.96 & 0.49 & 0.93 & 0.94 & 0.68 \\ 
   & DISTS~\cite{DBLP:journals/corr/abs-2004-07728} & 0.98 & 0.99 & \textbf{0.37} & 0.96 & 0.95 & 0.56 & \textbf{0.98} & 0.95 & \textbf{0.33} & 0.95 & 0.95 & 0.56 & 0.94 & 0.94 & 0.64 \\ 
   & MDSI~\cite{DBLP:journals/corr/NafchiSHC16} & 0.93 & 0.92 & 0.75 & 0.97 & 0.96 & 0.53 & 0.97 & 0.96 & 0.38 & 0.97 & 0.96 & {0.42} & 0.94 & 0.94 & 0.63 \\ 
   \midrule
   \parbox[t]{\tabcolsep}{\multirow{4}{*}{\rotatebox[origin=c]{90}{No Ref.}}} & BRISQUE~\cite{6272356} & 0.86 & 0.97 & 1.08 & 0.71 & 0.71 & 1.54 & 0.82 & 0.94 & 0.97 & 0.88 & 0.83 & 0.85 & 0.75 & 0.76 & 1.33 \\
 
   & MUSIQ~\cite{ke2021musiq} & 0.97 & 0.98 & 0.46 & 0.91 & 0.90 & 0.82 & 0.84 & 0.85 & 0.61 & 0.89 & 0.90 & 0.90 & 0.92 & 0.92 & 0.79 \\

 & MC360IQA~\cite{8702664}& 0.97 & 0.97 & 0.53 & 0.91 & 0.91 & 0.88 & 0.97 & \textbf{0.97} & 0.40 & 0.96 & 0.98 & 0.37 & 0.94 & 0.94 & 0.66 \\ 
   & VGCN*~\cite{xu2020blind} & 0.95 & 0.93 & 0.67 & \textbf{0.98} & 0.95 & 0.48 & \textbf{0.98} & 0.96 & 0.33 & \textbf{0.98} & 0.98 & \textbf{0.35} & 0.95 & 0.96 & 0.63 \\ 
   & VGCN+~\cite{xu2020blind} & 0.89 & 0.88 & 0.98 & 0.92 & 0.89 & 0.90 & 0.90 & 0.85 & 0.78 & 0.96 & 0.94 & 0.47 & 0.88 & 0.89 & 0.92 \\ 
   & ST360IQ (Ours) & \textbf{0.99} & \textbf{0.99} & {0.39} & 0.97 & \textbf{0.97} & 0.47 & 0.89 & 0.83 & 0.49 & 0.97 & \textbf{0.99} & 0.47 & \textbf{0.96} & \textbf{0.97} & \textbf{0.57} \\
\bottomrule
\multicolumn{8}{l}{\small{VGCN+ stands for the model trained with the same settings as our proposed method}} &
\multicolumn{9}{r}{\small{VGCN* stands for the results given in the original VGCN paper~\cite{xu2020blind}}}
\end{tabular}
}
\label{table:oiqa}
\end{table*}

\begin{table}[!t]
\caption{Contribution of using tangent viewports and saliency-guided sampling module to the final performance.}
\centering
\resizebox{0.98\linewidth}{!}{%
\begin{tabular}{lc@{$\;\,$}c@{$\quad$}c@{$\;\,$}c}
\toprule
& \multicolumn{2}{c}{CVIQ} & \multicolumn{2}{c}{OIQA} \\ 
        Method & PLCC$\uparrow$ & SRCC$\uparrow$  & PLCC$\uparrow$ & SRCC$\uparrow$\\ 
\midrule
Proposed model & 0.98 & 0.98 & 0.96 & 0.97\\ 

w/o saliency-guided sampling & 0.96 & 0.95 & 0.93 & 0.93\\ 
w/o tangent viewports  & 0.94 & 0.92 & 0.93 & 0.93 \\ 
\bottomrule
\end{tabular}
}
\label{table:ablation1}
\end{table}

\noindent\textbf{Performance Comparison.} We evaluate the performance of our model with three widely used metrics, namely Spearman's Rank Correlation Coefficient (SRCC), Pearson Correlation Coefficient (PLCC), and Root Mean Squared Error (RMSE). As commonly done, before computing PLCC and RMSE, we apply a five-parameter logistic function to the model's prediction. In Table~\ref{table:cviq}-\ref{table:oiqa}, we present the results of our model along with some traditional and learning-based metrics on the OIQA and CVIQ datasets. Most of these metrics are FR metrics and require an additional reference image. The remaining ones are NR metrics, within which MUSIQ~\cite{ke2021musiq} is the ViT-based model that we build our model on top of, and MC360IQA~\cite{8702664} and VGCN~\cite{xu2020blind} are the NR models specifically designed for omnidirectional images. 

Our proposed ST360IQ model, in general, outperforms the state of the art models in predicting $360^\circ$ image quality on all tested images. Training and applying MUSIQ on the ERP images give worse results than most of the NR metrics including ours as ERP  images suffers from additional spherical distortions, showing processing omnidirectional images needs special care, as we did in our proposed framework. Moreover, it is important to emphasize that for achieving these performances, our model does not require any pre-training involving IQA datasets that contain natural images. On the other hand, the highly competitive VGCN~\cite{xu2020blind} does, which can be also partly seen in Table~\ref{table:oiqa} in that its performance decreases without proper pretraining (cf. VGCN* and VGCN+).

\noindent\textbf{Effect of Sampling Module.} The performance of ST360IQ strongly relies on the utilized sampling strategies. We sample tangent viewports from a given spherical 360$^\circ$ image, and estimate its quality score from these viewports, instead of processing the whole image. Moreover, motivated by the human attention mechanism, we consider a saliency-guided sampling scheme to select the tangent viewports. We evaluate the contribution of these strategies in Table~\ref{table:ablation1}. First, we replace the tangent images with the image regions directly cropped from the salient parts of the corresponding ERP images. This leads to a significant performance loss, validating our claim that inherent distortions in ERP images have a negative influence on the quality predictions. Second, instead of selecting salient viewports, we perform random sampling over the spherical image and extract viewports accordingly regardless of their salience. Again, we observe that such kind of strategy gives poor performance as compared to our full model.

\begin{table}[!t]
\caption{Effect of using different embeddings on the performance within the proposed ST360IQ model.}
\centering
\resizebox{0.98\linewidth}{!}{%
\begin{tabular}{lc@{$\;\,$}c@{$\quad$}c@{$\;\,$}c}
\toprule
& \multicolumn{2}{c}{CVIQ} & \multicolumn{2}{c}{OIQA} \\ 
        Method & PLCC$\uparrow$ & SRCC$\uparrow$  & PLCC$\uparrow$ & SRCC$\uparrow$ \\ 
\midrule
Proposed model & {0.98} & {0.98} & {0.96} & {0.97} \\ 
w/o source embed. & 0.96 & 0.96 & 0.95 & 0.96\\ 
w/o geometric+source embed. & 0.94 & 0.95 & 0.93 & 0.94  \\ 
\bottomrule
\end{tabular}
}
\label{table:ablation2}
\end{table}

\noindent\textbf{Effect of geometric and source embeddings.} We assess the effect of using source and geometric embeddings on ST360IQ's performance by excluding them from the patch encodings. The results given in Table~\ref{table:ablation2} shows that adding geometric embeddings introduces significant performance gains. We conjecture that it allows the network to learn viewport-specific biases. Similarly, including source embeddings, in a way, let the model be aware of image-specific characteristics during training over the extracted salient viewports, further improving the prediction accuracies.

\section{Conclusion}
We propose a spherical-ViT based no-reference omnidirectional IQA method called ST360IQ, which predicts the quality score of a 360$^\circ$ image by  processing the image by extracting the most salient viewports, and aggregating the local quality scores estimated from them. Using the ViT-architecture allows us to better model the geometry of the spherical structure and the viewport biases. The effectiveness of the suggested approach is demonstrated by the experiments on two common 360$^\circ$ IQA datasets, which reveal that our model consistently attains the state-of-the-art performance.

\section{Acknowledgments}
This work was supported in part by KUIS AI Center Research Award, TUBITAK-1001 Program Award No. 120E501, GEBIP 2018 Award of the Turkish Academy of Sciences to E. Erdem, and BAGEP 2021 Award of the Science Academy to A. Erdem.

\clearpage
\bibliographystyle{IEEEbib}

\small{
\bibliography{example_report}}
\end{document}